\documentclass[10pt, a4paper]{article}

\usepackage[]{lrec-coling2024} 

\usepackage{amssymb}
\usepackage{kotex}
\usepackage{multirow}
\usepackage{graphicx}
\usepackage{longtable}
\usepackage{multicol}
\usepackage[normalem]{ulem}
\useunder{\uline}{\ul}{}
\usepackage[symbol]{footmisc}

\title{Hierarchical Graph Convolutional Network Approach for Detecting Low-Quality Documents}

\name{Jaeyoung Lee$^1$, Joonwon Jang$^2$, Misuk Kim$^{1,*}$\thanks{* Corresponding Author}} 

\address{$^1$Hanyang University, $^2$POSTECH \\
\{nowzer0, misukkim\}@hanyang.ac.kr \\
kaoara@postech.ac.kr}

\abstract{
Consistency within a document is a crucial feature indicative of its quality. Recently, within the vast amount of information produced across various media, there exists a significant number of low-quality documents that either lack internal consistency or contain content utterly unrelated to their headlines. Such low-quality documents induce fatigue in readers and undermine the credibility of the media source that provided them. Consequently, research to automatically detect these low-quality documents based on natural language processing is imperative. In this study, we introduce a hierarchical graph convolutional network (HGCN) that can detect internal inconsistencies within a document and incongruences between the title and body. Moreover, we constructed the \textit{Inconsistency Dataset}, leveraging published news data and its meta-data, to train our model to detect document inconsistencies. Experimental results demonstrated that the HGCN achieved superior performance with an accuracy of 91.20\% on our constructed \textit{Inconsistency Dataset}, outperforming other comparative models. Additionally, on the publicly available incongruent-related dataset, the proposed methodology demonstrated a performance of 92.00\%, validating its general applicability. Finally, an ablation study further confirmed the significant impact of meta-data utilization on performance enhancement. We anticipate that our model can be universally applied to detect and filter low-quality documents in the real world.
 \\ \newline \Keywords{Pre-trained language model, Detecting low-quality documents, Hierarchical graph convolutional network} }

\begin{document}

\maketitleabstract

\section{Introduction}
\label{sec:Intro}
In writing documents aimed at conveying information on specific subjects or purposes, such as research papers, essays, or news articles, it is crucial to write in alignment with the topics to convey precise meaning to the readers. The text must be written consistently, enhancing readability and making it easier for the reader to understand the author's intent \citep{garing2014coherence}. Therefore, consistency is essential in conveying the purpose and meaning to the readers and is a significant feature of text quality \citep{putra2017evaluating, xu2019cross}. While there is no uniform definition of consistency, a text can be considered consistent when it provides meaning to the readers \citep{wolf2005representing, feng2014impact}, is appropriately arranged around a particular theme for overall integration \citep{garing2014coherence}, or when all sentences are interrelated \citep{somasundaran2014lexical}.

Recently, with the massive generation of text across various media and online platforms, most people easily access desired information over a broader scope through these channels. However, a plethora of data from these media often creates confusion among readers due to low-quality documents that are inconsistent or irrelevant to the subject. Notably, readers usually approach documents based on their titles, exposing themselves to low-quality documents with incongruent headers and bodies or documents that initially contain relevant content but later include inconsistent advertising material. Such low-quality documents significantly fatigue the readers. Consequently, recent research employing natural language processing (NLP) techniques have been defined as various tasks to automatically detect multiple forms of low-quality documents \citep{abdolahi2016overview}.

Typically, the issue of mismatched titles and content in documents has been studied as the task of incongruent headline detection, referred to as clickbait \citep{jang2022detecting, mishra2020musem, mishra2021poshan, yoon2019detecting, yoon2021learning}. At the same time, tasks related to the consistency of documents have been researched under various tasks such as document discrimination, paragraph reconstruction, sentence insertion, summary coherence rating, sentence intrusion detection, etc \citep{shen2021evaluating}. However, traditional approaches to detecting low-quality documents, which usually focus on a single problem, such as incongruence between the headline and body or consistency within the document, are limited in their application to various types of low-quality documents. For example, datasets built for incongruent headline detection \citep{jang2022detecting, yoon2019detecting, yoon2021learning} only deal with incongruence between headline and body and do not consider consistency between sentences. Further, datasets for detecting inconsistency within the body \citep{jung2022detecting} do not address the problem of incongruent headline detection and are limited in that they use only the body within the dataset and do not utilize other information within the document. In actual data, there exists a variety of low-quality documents. Therefore, a model that detects only one type of low-quality document has limited applicability. There is a need for a universal model that can detect various types of low-quality documents.

We propose an adaptable model called hierarchical graph convolutional network (HGCN) that uses various document meta-data to detect two types of low-quality documents. HGCN uses meta-data to understand the consistency of the document, enabling the detection of incongruency between the headline and body. For the experimental verification of the proposed model, we have constructed an \textit{Inconsistency Dataset} consisting of 216,512 documents using actual published news articles, assuming inconsistency as a sentence intrusion situation as defined in \citet{jung2022detecting} and \citet{shen2021evaluating}. We performed a comparative analysis of the detection performance of our proposed HGCN model using the \textit{Inconsistency Dataset} constructed in this paper and the incongruent news headline dataset created by \citet{jang2022detecting}. Additionally, we conducted an ablation study to confirm the effectiveness of each sub-module applied to the HGCN model. All of our code and datasets will be publicly available on \href{https://github.com/nowzer0/HGCN-for-Detecting-Low-Quality-Documents}{https://github.com/nowzer0/HGCN-for-Detecting-Low-Quality-Documents} and are permitted for research purposes. The contributions of this research are as follows:

\begin{itemize}
\item In this research, an \textit{Inconsistency Dataset} comprising 216,512 documents, including news \texttt{title}, \texttt{subheading}, \texttt{body\_text}, and \texttt{image\_caption}, was constructed using real news articles.
\item We propose a universal hierarchical graph convolutional network (HGCN) that uses meta-data to detect consistency within the document and incongruency between the headline and body.
\item HGCN achieved significant performance in document inconsistency and incongruency between the headline and body, confirming its ability.
\end{itemize}

The subsequent sections will cover related work pertinent to this paper in Section \ref{sec:related_works}, describe the data used in this paper and the process of constructing the dataset in Section \ref{sec:dataset}, present the methodology proposed in this paper in Section \ref{sec:method}, report the experimental results using the constructed dataset in Section \ref{sec:Exp}, and present an ablation study to validate the effectiveness of the proposed model. Section \ref{sec:conclusion} discusses conclusions and future research.

\section{Related Works}
\label{sec:related_works}
Research on text coherence has been conducted with various definitions and across diverse task problems, which could lead to confusion regarding its meaning. To avoid such confusion, we follow the three definitions of similar document coherence proposed by \citet{jung2022detecting}. They categorize document coherence into coherence, consistency, and congruence. The coherence task involves detecting the flow of meaning by shuffling the sentence order within a document, producing incoherent documents, and assessing their incoherence. The consistency task defines documents as inconsistent when text from another document intrudes and evaluates such inconsistencies. The congruence task evaluates congruence between different types of texts, such as titles, bodies, and paragraphs. This paper aims to detect low-quality documents with inconsistency issues and incongruence between the headline and body. This section introduces existing research on inconsistency and incongruence detection and the related datasets.

\subsection{Inconsistency Detection}
\citet{shen2021evaluating} proposed the sentence intrusion detection task, which identifies a document's consistency by simulating an environment where sentences from external sources intrude. They constructed a dataset based on Wikipedia and CNN news articles and compared the performance of pre-trained language models on this task. \citet{jung2022detecting} proposed a graph-based document inconsistency detection model, treating each sentence as a node and the entire document as a node network. They defined the inconsistency detection task in the context of the sentence intrusion detection task proposed by \citet{shen2021evaluating} and constructed an inconsistency detection dataset by swapping sentences from news articles within the same category. However, their work was limited as they only used the body of news articles and did not utilize other critical information like titles, subheadings, and captions. Despite many low-quality documents being maliciously produced in the real world under sentence intrusion scenarios, related research is limited.

\subsection{Incongruent Detection}
Research on incongruent detection is more active than inconsistency detection. Typically, incongruent detection studies evaluate congruence by comparing the relationship between various text types, such as titles, subheadings, and bodies. Initial studies in incongruent detection employed linguistic and statistical features. For example, \citet{chen2015misleading} identified the likelihood of clickbait articles where the headline and body were misleading using linguistic and non-linguistic pattern analysis. \citet{chakraborty2016stop} manually constructed an incongruent dataset and detected incongruence using a support vector machine (SVM) based on linguistic and statistical features. \citet{wei2017learning} utilized features like the similarity between the title and body and applied a self-supervised method to use a large unlabeled dataset. Apart from methods that use linguistic and statistical features, deep learning models have also been employed to enhance the performance of incongruent detection. For instance, \citet{omidvar2019using} used a bi-directional gated recurrent unit (GRU) model to identify incongruence in terms of mean squared error (MSE). \citet{singhania20173han} proposed a three-tiered attention network to focus on essential words and sentences within the body and identify their relationship with the title. \citet{yoon2019detecting} employed a hierarchical network of GRUs to determine the relationship between the title and body. Furthermore, \citet{mishra2020musem} introduced a mutual attentive semantic matching model to reflect long body effectively and handle the length difference between titles and bodies. \citet{mishra2021poshan} proposed a POS-tag pattern-based hierarchical attention network using part-of-speech (POS) triplets and phrases. \citet{jang2022detecting} constructed an incongruent news headline dataset that includes metadata and introduced an attention network that utilizes additional text information. Recent research, such as \citet{yoon2021learning} and \citet{haque2023incongruity}, has encoded different types of text in documents, like titles and paragraphs, as nodes and detected incongruence using graph structures. However, the models proposed in these studies solely address the incongruent detection issue and are limited in addressing the inconsistency detection problem. Therefore, this paper aims to propose a model that universally detects low-quality documents encompassing both inconsistency and incongruent issues.

\subsection{Related Dataset}
\label{related_dataset}
According to \citet{yoon2021learning} and \citet{jang2022detecting}, the most significant challenges in detecting documents with mismatched titles and bodies, or inconsistency within the body, are (1) a lack of large datasets and (2) the absence of automated data preparation pipelines. To address these challenges, recent research has developed datasets with headline-body incongruence and body inconsistency through techniques like inserting or swapping the entire body of a document with similar subjects into the original document \citep{yoon2019detecting, yoon2021learning, jang2022detecting} or exchanging sentences within the body between document pairs \citep{jung2022detecting}. \citet{yoon2019detecting} extracted two news articles and created mismatched news by inserting the body content of one news article into the body of the other. However, this approach leads to incongruent news being longer than congruent news, and machine learning models may rely on text length rather than understanding the relationship between the headline and body, resulting in limitations. To solve this, \citet{yoon2021learning} extracted two similar news articles and swapped paragraphs to create a dataset without bias in length between incongruent and congruent data. They employed a pre-trained FastText\citep{bojanowski2017enriching} passed through the headline to measure Euclidean distance for similar news pair extraction. Still, this method is limited depending on the representation that pre-trained FastText captures. \citet{jang2022detecting} used a professional news outlet's `class code' feature, manually labeled by journalists for classification, to select two similar news articles and create incongruent news by swapping the entire body, satisfying conditions like transmission time to prevent false negatives. \citet{jung2022detecting} generated inconsistency news using the body only by swapping sentences in the same location in the body of two similar news articles selected using `class code'. We aim to build a more advanced dataset to identify document inconsistencies, building on the directions suggested by these prior studies.

\section{Dataset}
\label{sec:dataset}
\subsection{Data Description}
\label{data_description}
Using meta-data to develop a model for detecting low-quality documents with inconsistency issues, we constructed an \textit{Inconsistency Dataset}. The data used in the dataset construction covered various news fields and were obtained from one of South Korea's largest news outlets, Yonhap News. Yonhap News articles undergo strict verification procedures, with inappropriate ones deleted or revised, ensuring high credibility. The data spans 22 months of news articles published from January 2019 to October 2020 and consists of detailed information like the \texttt{title}, \texttt{subheading}, \texttt{body\_text}, \texttt{image\_caption}, category, class code, desk code, and send date. Class codes refer to specific categories like `Education,' `Economy,' and `Sports,' while the desk code indicates the author's department, and the send date is when the writer sent the manuscript to the editor. News articles with insufficient text information were excluded, such as photo news, photo essays, video news, infographics, news with too few class codes, or inappropriate news like breaking news without content in the body. Preprocessing was performed to remove various types of noise, including tags, URLs, Unicode like u+0000, and personal information, like the author's department, name, and email, from the text data within the news articles.

\subsection{Dataset Generation}
\begin{figure}[t!]
    \centering
    \includegraphics[scale=0.31]{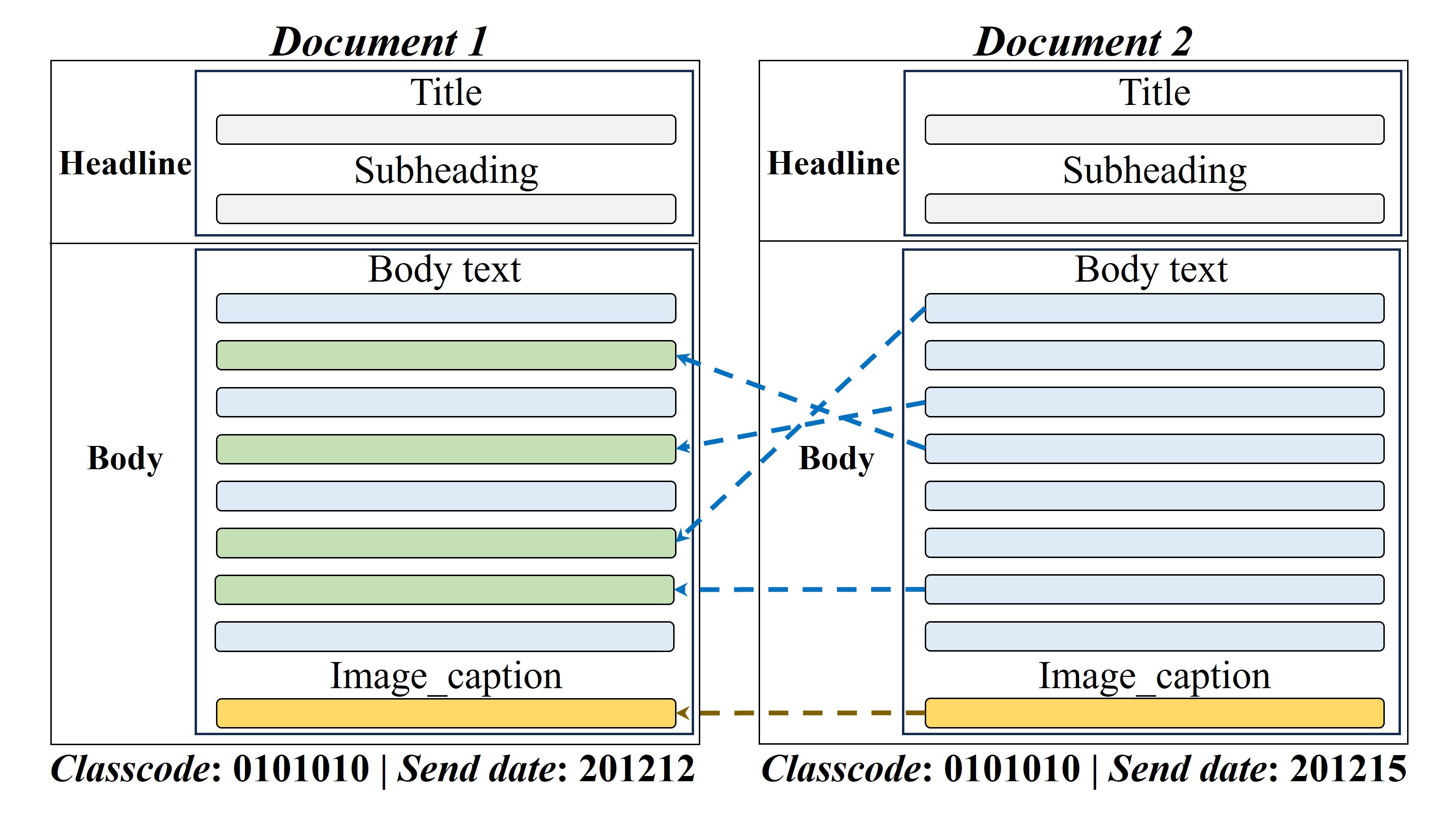}
    \caption{An example of the process to construct a dataset for detecting document inconsistencies.}
    \label{fig:datageneration}
\end{figure}

This research has developed a more advanced dataset to identify document inconsistencies, building on the datasets introduced in Section \ref{related_dataset}. Firstly, we follow the problem situations presented by \citet{jang2022detecting} and \citet{jung2022detecting}. Each news article has a headline, including the \texttt{title} and \texttt{subheading}, and a body comprising \texttt{the body\_text} and \texttt{image\_caption}. The \texttt{title}, \texttt{body\_text}, and meta-data, such as \texttt{subheadings} and \texttt{image\_captions}, are utilized during dataset construction. Each news article is marked with a label indicating its inconsistency status, and the model takes the \texttt{title}, \texttt{subheading}, \texttt{body\_text}, and \texttt{image\_caption} as input to detect inconsistent documents in a binary classification format. 
We have constructed an \textit{Inconsistency Dataset} using the preprocessed data described in Section \ref{data_description}. Figure \ref{fig:datageneration} illustrates the dataset generation process.
The construction of the dataset is carried out through an automated process consisting of three steps using the `class code': (1) sampling target articles from the entire set of articles, (2) selecting article pairs with similar body lengths within the same class code, and (3) swapping sentences within the body of the article. First, all articles are divided into two categories: consistent and inconsistent. The process continues with the only inconsistent data, where a pair of articles is selected to build inconsistent data. For a challenging dataset construction, three rules were established: first, articles within the same class code were used to prevent the insertion of heterogeneous content. Since only articles from the same class are utilized, the model requires a more profound understanding to address the issue. Second, articles with similar body lengths were chosen to ensure the model does not rely on text length for its solutions. Lastly, overlapping send dates are avoided to prevent false negatives. Following this, a selected pair of articles have a set shuffle ratio of their sentences swapped to create inconsistent data. The shuffle ratio is set to 0.5, and through this automated construction process, a dataset is created that requires the identification of the consistency of documents.

\section{Proposed Method}
\label{sec:method}
\begin{figure*}[ht!]
    \centering
    \includegraphics[width=0.9\textwidth]{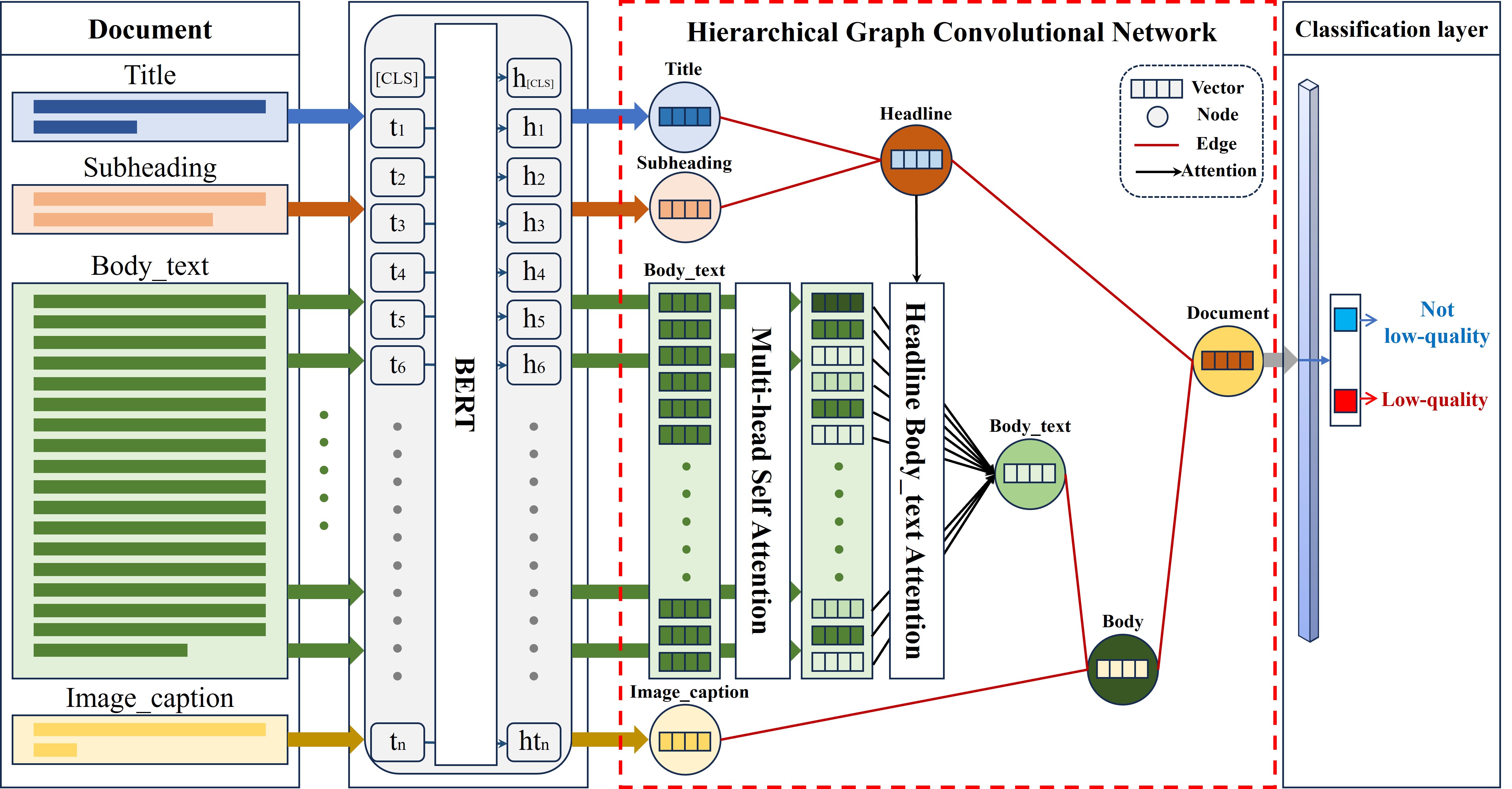}
    \caption{An overview of the hierarchical graph convolutional network (HGCN). Each document component is encoded using BERT to vectorize the components. The components are assembled as nodes undergoing the graph convolutional layer process. The document vector processed through the graph convolutional layer is finally passed to the classification layer to detect low-quality documents.}
    \label{fig:model}
\end{figure*}

We constructed a graph reflecting the document's hierarchical structure, enabling the model to directly comprehend the relationships between the various components. The overall structure of the HGCN model is illustrated in Figure \ref{fig:model}. 
As shown in Figure \ref{fig:model}, leaf nodes consist of \texttt{title}, \texttt{subheading}, \texttt{body\_text}, and \texttt{image\_caption} vector nodes, while the root node comprises a document vector node, reflecting the hierarchical structure of the document. Edges were placed between related nodes to allow the propagation of information. The HGCN model was trained to minimize the following loss function in Eq.(\ref{eq1}):
\begin{equation}\label{eq1}
    \small L = -\frac{1}{N}\sum_{i=1}^{N}\left[y_{i} \cdot log(\hat{p_{l}}) + (1 - y_{i} \cdot log(1 - \hat{p_{l}})\right].
\end{equation}
$N$ represents the total number of data, $y$ is a label signifying the presence or absence of consistency in document $i$, and $p$ is the probability that the document is inconsistent, as predicted by the classification model in Section \ref{hgcn}. The components within the model are detailed in Sections \ref{sentence_embedding} through \ref{Method_MLP}.

\subsection{Sentence Embedding}
\label{sentence_embedding}
To vectorize the document's \texttt{title}, \texttt{subheading}, \texttt{body\_text}, and \texttt{image\_caption}, we employed Bidirectional Encoder Representations from Transformers (BERT) \citep{devlin2019bert}. Since the \texttt{title}, \texttt{subheading}, and \texttt{image\_caption} are mostly less than 512 tokens, they were encoded as a single sentence vector, while the multi-sentenced \texttt{body\_text} was encoded sentence by sentence. Among various encoding methods, such as using the last layer's [CLS] token vector, averaging the hidden representations of each token, or concatenating the top four hidden representations of the BERT encoder's [CLS] tokens, the highest performance was achieved using the last layer's [CLS] token as a contextual representation. This approach was employed as sentence embedding in this paper. The document's \texttt{title}, \texttt{subheading}, \texttt{body\_text}, and \texttt{image\_caption} were padded or truncated to the experimentally defined hidden dimension $m$, with $k=1$ for the \texttt{title}, \texttt{subheading}, \texttt{image\_caption}, and $k$ set to the 80th percentile for the number of sentences in the \texttt{body\_text} of the entire document. In this study, we set $k$ to 21.

\subsection{Headline Representation}
\label{headline_rep}
We integrated the \texttt{title} and \texttt{subheading} into a headline to reflect the document's hierarchical structure. Initially, the \texttt{title} and \texttt{subheading} vectors were concatenated, then passed through a 2-layer multi-layer perceptron (MLP) consisting of a linear layer and non-linear function to compute the headline representation. The equation for calculating the headline representation is given in Eq.(\ref{eq2}).
\begin{equation}\label{eq2}
    e_{headline} = MLP([e_{title}; e_{subheading}]),
\end{equation}
Where $e_j\in \mathbb{R}^{m \times 1}$ denotes the sentence embedding vector of the $j$ component, and the Rectified Linear Unit (ReLU) function was used as the non-linear function. A down projection was applied in the second MLP layer to ensure that the dimension of $e_{headline}$ was consistent with that of $e_{title}$ and $e_{subheading}$.

\subsection{Body Representation}
\label{body_rep}
We employed self-attention to reflect the relationships between each sentence in the \texttt{body\_text}. This is because self-attention can yield a context vector with strengthened relationships between the congruent sentences related to the headline, which is advantageous for detecting inherent inconsistency in the \texttt{body\_text}. The sentence vector of the \texttt{body\_text} was calculated to produce a context-inclusive representation by passing it through a multi-head self-attention layer with eight heads. Self-attention output of the \texttt{body\_text} is defined as $S_{inconsistency} \in \mathbb{R}^{m\times k}$.

According to \citet{yoon2021learning}, understanding the relationship between the headline and body is vital for effectively detecting articles with incongruence between the headline and body. By dot-producting $e_{headline}$ with \texttt{body\_text} self-attention $A_{body\_text}$ as in Eq.(\ref{eq3}), one can obtain a $s_{incongruent}$ vector that reveals the relationship between sentences of the headline and the \texttt{body\_text}.
\begin{equation}\label{eq3}
    \small s_{incongruent} = softmax(e_{headline}^{T}\cdot A_{body\_text})^T,
\end{equation}
where $s_{incongruent}$ $\in$ $\mathbb{R}^{k \times 1}$ represents the vector indicating consistency scores between each headline sentence and \texttt{body\_text}. The quality of \texttt{body\_text} that reflects consistency information between sentences of \texttt{body\_text} from $S_{inconsistency}$ and congruence information between headline-\texttt{body\_text} from Eq.(\ref{eq3}) is given as $s_{body\_text}$ $\in$ $\mathbb{R}^{k \times 1}$ and can be computed as in Eq.(\ref{eq4}).
\begin{equation}\label{eq4}
    \resizebox{2.7in}{0.11in}{$s_{body\_text} = \alpha \cdot s_{incongruent} + (1-\alpha) \cdot (w^T\cdot S_{inconsistency})^T,$}
\end{equation}
where $\alpha$ determines the weight of congruence and consistency reflection, and $w \in \mathbb{R}^{m \times 1}$ is a uniform vector for linear combination. Finally, the embedding vector $e_{body\_text}$ of \texttt{body\_text} was calculated by multiplying the score from Eq.(\ref{eq4}) with the self-attention from $S_{inconsistency}$ as shown in Eq.(\ref{eq5}).
\begin{equation}\label{eq5}
    e_{body\_text} = S_{inconsistency} \cdot s_{body\_text}.
\end{equation}
Similar to the headline, to reflect the hierarchical nature, we concatenated \texttt{body\_text} and \texttt{image\_caption} to represent and utilize it as a body vector, as shown in Eq.(\ref{eq6}).
\begin{equation}\label{eq6}
    e_{body}=MLP([e_{body\_text};e_{image\_caption}]),
\end{equation}
where $MLP$ utilizes two layers composed of a linear layer and a non-linear function. As in Section \ref{headline_rep}, the ReLU function was employed as the non-linear function, and a down-projection was applied in the second layer for the convenience of subsequent computations.

\subsection{Document Representation}
\label{Method_MLP}
Using the headline and body vector calculated from Section \ref{headline_rep} and Section \ref{body_rep}, the representation vector that encompasses the overall meaning of the document was computed as in Eq.(\ref{eq7}).
\begin{equation}\label{eq7}
    e_{document} = MLP([e_{headline};e_{body\_text}]).
\end{equation}
The final document representation was computed by concatenating the headline and body vectors as in Eq.(\ref{eq7}) and using them as the input for a 2-layer MLP composed of a linear layer and a non-linear function. This model structure allows us to capture the hierarchical features of the document.

\subsection{Hierarchical Graph Convolutional Model}
\label{hgcn}
We propagated information between nodes using the graph convolution layer. Specifically, we performed a linear transformation by multiplying the weights to each node. Subsequently, a non-linear function and skip-connection were applied to prevent the over-smoothing problem. We applied the graph convolution layer $n$ times, and the transformation from the $l^{th}$ layer to the $(l+1)^{th}$ layer is as in Eq.(\ref{eq9}).
\begin{equation}
\begin{array}{cc}
    D^{(l+1)} = \sigma (\tilde{D}^{-\frac{1}{2}}\tilde{A}\tilde{D}^{-\frac{1}{2}}D^{(l)}W^{(l)}),\\
    D^{(l+1)} = D^{(l+1)} + D^{(l)},
\end{array}
\label{eq9}
\end{equation}
where $D$ is the document matrix concatenated sequentially from the \texttt{title}, \texttt{subheading}, \texttt{body\_text}, \texttt{image\_caption}, \texttt{headline}, \texttt{body}, and \texttt{document} vectors. $\tilde{D}$ is a matrix indicating the degree of the graph, and $\tilde{A}$ represents the adjacency matrix. We utilized an undirected graph\cite{kipf2017semisupervised}. One can refer to Figure \ref{fig:model} to check which nodes are connected to which nodes. $W^{(l)}$ is the learning weight for the $l^{th}$ layer, a matrix for linear transformation. The ReLU function was used as the non-linear function. After performing the hierarchical graph convolutional layer $n$ times, the document representation was passed through the final 2-layer of MLP for binary classification to detect the inconsistency of the document. In this study, the graph convolutional layer was set to 4 through hyperparameter search, and thus it was performed over $n=4$ time.

\section{Experimental Results}
\label{sec:Exp}
\subsection{Experimental Setup}
Model construction and the overall training process were conducted through the PyTorch framework. The hyperparameters used in the HGCN model experiment were set as follows: batch size 92, learning rate 1e-05, AdamW optimizer, hidden dimension 768, dropout rates 0.1, additional hyperparameters alpha 0.3, and GCN layer count 4 for reflecting the document's headline and \texttt{body\_text} representation. Experiments on the alpha and GCN layer settings are detailed in section \ref{Ablation_study}. The dataset used for training and experimentation consists of 216,512 samples, with a train-valid-test ratio of 9:0.5:0.5, resulting in 194,860 training data, 10,826 validation data, and 10,826 test data. The characteristics of the dataset are shown in Table \ref{stat_of_dataset}.
\begin{table}[h!]
\centering
\resizebox{\columnwidth}{!}{%
\begin{tabular}{cccccccc}
\hline
              & Mean  & Stdev & Min & Max & 1Q & 3Q  & \begin{tabular}[c]{@{}c@{}}Number \\ of tokens\end{tabular} \\ \hline
\texttt{Title}         & 1.00     & 0.07  & 1   & 4   & 1    & 1    & 29.09                  \\ 
\texttt{Body\_text}     & 14.64 & 8.97  & 1   & 138 & 8    & 19   & 828.07                 \\ 
\texttt{Subheading}    & 1.00     & 0.05  & 0   & 4   & 1    & 1    & 42.48                  \\ 
\texttt{Image\_caption} & 1.23  & 0.69  & 0   & 15  & 1    & 1    & 45.51                  \\ \hline
\end{tabular}%
}
\caption{Characteristics of the Inconsistency Dataset.}
\label{stat_of_dataset}
\end{table}
The mean, standard deviation (Stdev), minimum (Min), maximum (Max), 25\% quantile (1Q), and 75\% quantile (3Q) values are related to the number of sentences. Meanwhile, the last column, number of tokens, represents the average number of tokens per element.
All experiments were conducted in an environment with Intel(R) Xeon(R) Silver 4210R CPU @ 2.40 GHz and TITAN RTX 4ea.

\subsection{Comparative Models}
To evaluate the performance of the proposed model, a baseline that detects document inconsistency must be utilized, but research on models for detecting document inconsistency is limited. Therefore, comparative experiments with models showed good performance in detecting incongruencies between text types. However, some models only accept the \texttt{title} and \texttt{body\_text} as inputs. To ensure a fair comparison, the \texttt{subheading} and \texttt{image\_caption} were concatenated respectively to the end of the \texttt{title} and \texttt{body\_text} for use. The detailed explanations for each comparative model are as follows: \\
\textbf{TF-IDF+SVM \citep{chakraborty2016stop}}: Linguistic patterns from news headlines were extracted and detected using SVM. In this study, \texttt{title}, \texttt{subheading}, \texttt{body\_text}, and \texttt{image\_caption} were concatenated, and features were extracted through Term frequency-Inverse document frequency (TF-IDF) to use an SVM model for comparison. \\
\textbf{GRU-Avg \citep{rashkin2017truth}}: All GRU hidden states were averaged to use as news representation. For a fair comparison with the proposed model, \texttt{title}, \texttt{subheading}, \texttt{body\_text}, and \texttt{image\_caption} were concatenated and passed through the word-embedding layer and GRU. \\
\textbf{GRU-Last \citep{rashkin2017truth}}: The last hidden state of the GRU was used as news representation. Like GRU-Avg, \texttt{title}, \texttt{subheading}, \texttt{body\_text}, and \texttt{image\_caption} were concatenated and passed through the word-embedding layer and GRU. \\
\textbf{3HAN \citep{singhania20173han}}: It is an attention network composed of three layers, and the representation after passing through them is utilized. The \texttt{body\_text} and \texttt{image\_caption} were concatenated and used as the body, and the \texttt{title} and \texttt{subheading} were concatenated to form the headline. \\
\textbf{AHDE \citep{yoon2019detecting}}: A hierarchical network formed with GRUs maps each sentence in the body into a one-dimensional vector. Attention is applied with the headline vector to weight important sentences. Like 3HAN, the \texttt{body\_text} and \texttt{image\_captions} were concatenated for the body, and the \texttt{title} and \texttt{subheading} were concatenated for the headline. \\
\textbf{Headline-Body Attention Model \citep{jang2022detecting}}: This is one of the state-of-the-art models for incongruent detection. For the first time, the authors presented a hierarchical attention network that utilizes additional text information such as \texttt{subheadings} and \texttt{image\_captions}. \\

\subsection{Experimental Results of Low-quality Document Detection}
This study conducted experiments on the dataset we constructed and the dataset released by \citet{jang2022detecting}. Since the distribution of labels within the dataset is balanced, accuracy was used as the evaluation metric. The experimental results can be found in Table \ref{Exp}.
\begin{table}[ht!]
\centering
\resizebox{\columnwidth}{!}{%
\begin{tabular}{l|cc}
\hline
\multicolumn{1}{c|}{\multirow{3}{*}{Model}} & \multicolumn{2}{c}{Accuracy}                        \\ \cline{2-3} 
 & \multicolumn{1}{c|}{\begin{tabular}[c]{@{}c@{}}Inconsistency \\ dataset\end{tabular}} & \begin{tabular}[c]{@{}c@{}}Incongruent \\ dataset\end{tabular} \\ \hline
TF-IDF + SVM            & \multicolumn{1}{c|}{73.96}         & 50.12          \\
GRU-AVG                  & \multicolumn{1}{c|}{64.76}         & 52.01          \\
GRU-LAST                    & \multicolumn{1}{c|}{60.41}         & 59.48          \\
3HAN                     & \multicolumn{1}{c|}{74.37}         & 75.25          \\
AHDE                     & \multicolumn{1}{c|}{78.12}         & 85.03          \\
Headline-Body Attention Model        & \multicolumn{1}{c|}{{\ul 83.12}}   & \textbf{93.68} \\ \hline
HGCN                                                 & \multicolumn{1}{c|}{\textbf{91.20$^\ddagger$}} & {\ul 92.00}    \\ \hline
\end{tabular}%
}
\caption{\label{Exp}Performance comparison(\%) of inconsistency and incongruent detection. $^\ddagger$ denotes
that the highest result is statistically significant at $p < 0.01$ compared to the second best, using a paired t-test.}
\end{table}

First, when comparing the experimental results of the \textit{Inconsistency Dataset} constructed in this study, the proposed model showed the highest performance of 91.20\%. This suggests that the multi-head self-attention helps understand relationships between sentences in the \texttt{body\_text}, allowing the representation of sentences to include inconsistencies. Additionally, it can be inferred that the model with a hierarchical structure, based on the representation of each sentence in the \texttt{body\_text} containing inconsistencies and the representation of the \texttt{body\_text} created by headline attention, effectively made a document representation through graph convolutional operations. On the other hand, models using GRU showed unsatisfactory performance, with GRU-AVG at 64.76\% and GRU-LAST at 60.41\%. This is likely because when concatenating the \texttt{title}, \texttt{subheading}, \texttt{body\_text}, and \texttt{image\_caption}, the resultant long length faces the long-term dependency problem. Moreover, the model that extracted linguistic features using TF-IDF and classified them via SVM showed a performance of 73.96\%. While TF-IDF-based features were more useful for classification than simple GRU-based models, they still showed limitations compared to neural network models based on hierarchical structures. The 3HAN showed a slightly higher performance of 74.37\% compared to the SVM. This could be because the same attention network was applied to every sentence in the \texttt{body\_text}, limiting the extraction of unique meaning from each sentence. AHDE and the Headline-Body Attention Model each achieved a performance of 78.12\% and 83.12\%, indicating that using attention to grasp the relationship between the body and the headline is beneficial in detecting inconsistencies between the headline and the body. However, models based on AHDE and attention have limitations in detecting inconsistencies between sentences within the \texttt{body\_text}.

\citet{jang2022detecting} released a dataset where the entire body content is unrelated to the headline while proposing an attention model that uses additional text information. Unlike the dataset proposed in our study, it does not contain inherent inconsistencies, and the authors mention that the headline-body attention they proposed played a significant role in performance improvement. To verify the robustness of the model proposed in our study, experiments were conducted on the dataset constructed for the incongruity detection task. The results can be found in the incongruent dataset column of Table \ref{Exp}. 
As a result of the experiment, the performance of the model proposed in this study was recorded at 92.00\%. This is 1.68\% lower than the model presented by \citet{jang2022detecting}. Since the model presented by \citet{jang2022detecting} reflects only the incongruence between the headline and the body, it performed well on that dataset. However, since it doesn't reflect inconsistencies within the document, its performance dropped significantly in cases of inconsistency, as shown in the inconsistency dataset column of Table \ref{Exp}. On the other hand, the proposed HGCN model shows sufficiently good performance even on the incongruent detection dataset. This suggests that the representation of each sentence in the \texttt{body\_text} containing inconsistencies and the \texttt{body\_text} representation created by headline attention significantly impact understanding the inherent incongruity in the data. These results confirm that our model can effectively detect inconsistency and incongruence, proving its robustness compared to existing models.

\subsection{Ablation Study}
\label{Ablation_study} 
\subsubsection{Effectiveness of Number of GCN Layers}
According to \citet{li2018deeper}, as the number of layers in a graph neural network increases, there is a heightened probability that nodes will possess overlapping receptive fields. This leads to a decrease in the expressive capability of the representation, resulting in the well-known over-smoothing problem. To resolve this, following the suggestions by \citet{oono2020graph} and \citet{huang2020tackling}, we strengthened the expressiveness by adding a 2-layer MLP for nonlinear transformation to the node transformation process in this study. Additionally, as the node embedding of previous graph neural network layers could contain meaningful information for classification, we added shortcuts between layers to conduct skip-connections. However, these previously mentioned methods did not completely solve the over-smoothing problem, so we compared the performance changes according to the number of layers, and the results are shown in Table \ref{ablation01}.
\begin{table}[h!]
\centering
\resizebox{\columnwidth}{!}{%
\begin{tabular}{cccccc}
\hline
GCN Layer & 1    & 2    & 3    & 4             & 5    \\ \hline
Accuracy  & 85.60 & 86.70 & 88.20 & \textbf{91.20} & 89.00 \\ \hline
\end{tabular}%
}
\caption{\label{ablation01}Effectiveness results based on the number of GCN layers.}
\end{table}
The performance increased as the number of layers increased due to the influence of MLP and skip-connection but started to decline after exceeding four layers. Therefore, all subsequent experiments were conducted with four layers.

\subsubsection{Effectiveness of Headline Attention}
The proposed model creates an advantageous representation for detecting the inherent inconsistency in the document through \texttt{body\_text}'s multi-head self-attention. Moreover, headline attention produces an easily interpretable representation of the incongruity between the headline and the body. Hence, the optimal harmony of these two pieces of information is essential to maximize the detection performance of the proposed model. We measured the model's performance changes according to the alpha value controlling the headline attention reflection ratio. The experimental results are shown in Table \ref{ablation02}.
\begin{table}[h!]
\centering
\resizebox{\columnwidth}{!}{%
\begin{tabular}{ccccccll}
\hline
Alpha    & 0.1  & 0.2  & 0.3           & 0.4  & 0.5  & 0.6  & 0.7  \\ \hline
Accuracy & 87.80 & 87.80 & \textbf{91.20} & 88.40 & 89.20 & 88.90 & 88.10 \\ \hline
\end{tabular}%
}
\caption{\label{ablation02}Effectiveness results of headline attention.} 
\end{table}
As shown in Table \ref{ablation02}, the best performance was observed when alpha was 0.3. Therefore, 0.3 was used as the reflection ratio of the headline presentation in the experiment.

\subsubsection{Significance of Meta-data Information}
Most existing models have typically used only the \texttt{title} and \texttt{body\_text}. Therefore, we evaluated the utility of \texttt{subheading} and \texttt{image\_caption}, and the experimental results are shown in Table \ref{ablation03}.
\begin{table}[h!]
\centering
\resizebox{\columnwidth}{!}{%
\begin{tabular}{ccccc}
\hline
\texttt{Title} & \texttt{Body\_text} & \texttt{Subheading} & \texttt{Image\_caption} & Accuracy       \\ \hline
$\vee$     & $\vee$    & $-$             & $-$       & 80.78          \\
$\vee$     & $\vee$    & $\vee$          & $-$       & 86.10          \\
$\vee$     & $\vee$    & $-$             & $\vee$    & 89.02          \\
$\vee$     & $\vee$    & $\vee$          & $\vee$    & \textbf{91.20} \\ \hline
\end{tabular}%
}
\caption{\label{ablation03}Significance results of meta-data information. ``$\vee$" indicates the use of meta-data, while ``$-$" signifies its non-use.}
\end{table}
Without utilizing both \texttt{subheading} and \texttt{image\_caption}, the model exhibited a performance of 80.78\%, which is about a 10\% decline, signifying that the additional information had a meaningful impact on node representation creation. When only the \texttt{image\_caption} was excluded, the performance was 86.10\%; when only the \texttt{subheading} was excluded, it was 89.02\%. This can be interpreted as a result of the \texttt{image\_caption} node propagating helpful information for detection during the process where the \texttt{body\_text} node, mixed with inconsistency and incongruent information, propagates information to the body node, unlike the \texttt{subheading} node that contributes to the creation of the headline representation node along with the \texttt{title} node. These experimental results prove that the proposed model operates effectively, reflecting the document's meta-data and hierarchical structure.

\subsection{Qualitative Analysis}
We have conducted a qualitative analysis of low-quality documents that our model misclassified. First, unlike the \texttt{title} and \texttt{body\_text}, some news articles lack a \texttt{subheading} or \texttt{image\_caption} or are composed of short phrases with limited information. The model showed a tendency to struggle to detect low-quality documents that are either incongruent or inconsistent, especially when there is no \texttt{subheading} or \texttt{image\_caption} or when they are briefly composed. Second, we encounter documents related to various domains. Among these, documents about specialized domains such as medicine, law, and accounting can be challenging to comprehend, even for humans, unless they are experts in the field. The model had difficulty detecting low-quality documents with complex content in such specialized domains. Lastly, there are cases where the content seems to cover the same issue as the \texttt{title} but subtly includes different content related to the same issue. There are low-quality documents that subtly swap contents within the \texttt{body\_text} to induce reader fatigue. 
These low-quality documents pose a particularly challenge for the model's detection capabilities due to their cunning inclusion of content that is superficially similar (for more details, see Appendix \ref{Appendix}).
To effectively classify these difficult-to-detect low-quality documents, the model needs to understand more information, such as sequences, numerics, dates, and domain knowledge and requires a higher level of contextual understanding. Additionally, it is anticipated that utilizing more metadata could improve detection performance.

\section{Conclusions}
\label{sec:conclusion}
To detect low-quality documents, we proposed an HGCN model that utilizes meta-data to identify inconsistencies within the document and incongruence between the headline and body. Furthermore, we constructed an \textit{Inconsistency Dataset} comprising 216,512 articles with \texttt{title}, \texttt{subheading}, \texttt{body\_text}, and \texttt{image\_caption} using data from Yonhap News, one of the largest news outlets in South Korea. In datasets related to inconsistency and incongruence, the HGCN showed performances of 91.20\% and 92.00\%, respectively, outperforming most comparative models. Furthermore, through various comparative experiments, we verified the utility of the proposed model's components and confirmed that the utilized meta-data significantly impacts performance. Our model can detect low-quality documents with inconsistency and incongruency issues between the headline and body. It is anticipated to be universally applicable for evaluating documents generated across diverse media, identifying and filtering out low-quality ones. 

Recent advancements in Large Language Models have demonstrated impressive performance across various NLP tasks. However, the substantial size of these models imposes limitations on their use. Nevertheless, an efficient two-phase approach can be employed, where initial inference is conducted using our resource-efficient model, followed by applying Large Language Models to detect samples not accurately classified in the first phase. This method has the potential to enhance the performance of low-quality document detection. In the future, we aim not only to detect low-quality documents with inconsistency and incongruence issues but also to expand the model structure and leverage more meta-data to detect a broader range of low-quality document types.

\section*{Acknowledgements}
This work was supported by the Technology Innovation Program (or Industrial Strategic Technology Development Program-Knowledge service industry technology development project - service core technology development) funded by the Ministry of Trade, Industry \& Energy (MOTIE, Korea) (Project Number: 20018758) and the National Research Foundation of Korea (NRF) grant funded by the Korea government (MSIT) (RS-2023-00247442).

\nocite{*}
\section*{Bibliographical References}
\bibliographystyle{lrec-coling2024-natbib}
\bibliography{lrec-coling2024-example}

\section*{Appendix}
\appendix
\setcounter{table}{0}
\renewcommand{\thetable}{\Alph{section}.\arabic{table}}
\section{Examples of Qualitative Analysis}
\label{Appendix}
We analyzed low-quality documents misclassified by our model. Some cases of misclassification are as follows, and examples are provided in Table \ref{appendix1}, \ref{appendix2}, and \ref{appendix3}. The examples in the table are translated articles written in Korean.

\textbf{Documents Lacking Metadata Information.}
Unlike the title and body text, some news articles lack a subtitle or image caption or are composed of short phrases with limited information. In Table \ref{appendix1}, the title covers the extension of tax exceptions in the agricultural sector in the Jeollanam-do region and its reflection in the government's tax law amendment. The body text, starting from the first sentence, discusses the need for the Ministry of Environment to permanently open the four major rivers and to open the Hapcheon-Changnyeong and Changnyeong-Haman weirs in the Gyeongsangnam-do region. This indicates a difference between the title and body text, and the body itself is an inconsistent, low-quality document. The model misclassified documents of low quality, demonstrating a tendency to struggle with detection in cases like Table \ref{appendix1}, where subtitles or image captions are absent or are composed briefly.

\textbf{Documents Related to Specialized Domains.}
We encounter documents related to various domains. Among these, documents about specialized domains such as medicine, law, and accounting can be challenging to comprehend, even for humans, unless they are experts in the field. In Table \ref{appendix2}, the title suggests content related to the expansion of mobile Hometax and the National Tax Service's Son-Tax for year-end tax settlement and issuance of cash receipts. This example is a low-quality document that, in the body text, combines content relevant to the title, such as tax filing and refunds in the tax domain, with unrelated content regarding the relaxation of regulations for contributions of shares to public interest corporations to activate SMEs' business succession. The model misclassified this low-quality document. In cases like this example, where the content pertains to complex topics in specialized domains, the model struggles with understanding, leading to misclassification.

\textbf{Documents Involving the Same Issue.}
There are low-quality documents that subtly swap contents within the body text to induce reader fatigue. Table \ref{appendix3} documents the opposition to pension reform in France. At first glance, the title and body text appear to cover the same topic, and the content of the body text seems consistent. However, it is a low-quality document created by swapping content between articles related to the third and fourth opposition rallies against French pension reform. Such low-quality documents are particularly challenging to detect because they cunningly include similar content. To detect low-quality documents like Table \ref{appendix3}, the model needs a higher level of contextual understanding, including grasping more information like the sequence and date of the events. Additionally, it is expected that incorporating more metadata could improve detection performance.
\begin{table*}[h!]
\centering
\begin{tabular}{c|p{12cm}}
\hline
\texttt{Title}         & Jeollanam-do to Extend Tax Exemptions in Agriculture, Reflect in Government's Tax Law Amendment \\ \hline
\texttt{Subheading}    & Null                                                                                            \\ \hline
\texttt{Body\_text} &
  Amid the Ministry of Environment's decision to dismantle or permanently open some of the four major river weirs, Gyeongsangnam-do has drawn attention by stating that the Hapcheon-Changnyeong and Changnyeong-Haman weirs in the region should be opened as much as possible. During the 361st temporary meeting of the Gyeongsangnam-do Assembly on the 6th, Lee Sang-yeol, a member of the Democratic Party of Korea in the Economic and Environmental Committee, stated, "Opening the gates of the Hapcheon-Changnyeong and Changnyeong-Haman weirs is expected to reduce algae." The content proposed by Jeollanam-do to the government in June of this year includes provisions specified in the Special Taxation Act, which is a national tax. These consist of six cases of direct support to farmers, such as applying a reduced VAT rate to agricultural materials supplied to farmers and tax exemption on interest income from farmers' savings, and five cases of indirect support to agricultural corporations, like tax exemption on deposits and contributions to agricultural cooperatives. The continuous tax exemption measures include interest income from farmers' savings and interest income from cooperative deposits up to 30 million won and dividend income from contributions up to 10 million won. The reduced VAT rate will continue to apply to agricultural materials supplied to farmers to reduce the burden of farming costs, and VAT exemption will also apply to agricultural materials directly imported by farmers. The low tax rate on net profit of cooperatives such as agricultural cooperatives and tax reduction for small and medium-sized enterprises engaged in crop cultivation and livestock farming will also be maintained. The article also asked about the province's position on fully opening these weirs. According to the revised proposal, any purchaser of rural housing, regardless of size, can benefit from a transfer income tax exemption, which currently only applies to purchasers of rural housing under 660㎡. Jeong Seok-won, Director of Environment and Forestry, answered, "The opening of the weirs is being operated flexibly to minimize the impact on the surrounding areas, such as intake and discharge facilities and agricultural damage." He added, "To improve the water quality of the Nakdong River, such as reducing algae, it is necessary to open the weirs as much as possible, but full opening should be preceded by sufficient monitoring and preparation of damage measures." He continued, "The province, in collaboration with five related departments such as the Agricultural Policy Department and cities and counties including Changwon, Euiryeong, Haman, Changnyeong, and Hapcheon, has formed a response team to check for damage due to the opening of the weirs." He also mentioned, "The Ministry of Environment plans to commission a service this month to minimize agricultural damage due to groundwater caused by the opening of the weirs and will establish damage measures based on the results." \\ \hline
\texttt{Image\_caption} & Gyeongsangnam-do Provincial Council Plenary Hall                                                \\ \hline
\end{tabular}
\caption{Examples of documents lacking metadata information.}
\label{appendix1}
\end{table*}

\begin{table*}[t]
\centering
\begin{tabular}{c|p{12cm}}
\hline
\texttt{Title}         & Year-end Tax Settlement and Issuance of Cash Receipts All Possible via Mobile Hometax \\ \hline
\texttt{Subheading}    & Nationwide Expansion of National Tax Service's Son-Tax Mobile Application             \\ \hline
\texttt{Body\_text} &
  The National Tax Service's mobile Hometax application, Son-Tax, is set to be expanded this year to the level of PC Hometax. There have been calls for the relaxation of regulations related to the contribution of shares to public interest corporations to activate the succession of small and medium-sized enterprises (SMEs). Researcher Kim Hee-sun from the SME Research Institute argued on the 6th in a report on utilizing public interest corporations and special stocks to promote the succession of SMEs, stating, "This is a realistic alternative that can be implemented relatively quickly within the existing system." All taxpayers can file taxes such as transfer, gift, consumption, withholding, comprehensive real estate, education, recognition, and liquor taxes on mobile, and VAT, comprehensive income, and securities transaction taxes are also possible, excluding some taxpayers. From the beginning of this year, wage earners can handle the entire year-end tax settlement process on mobile. Companies can also create, modify, and submit wage income statements on mobile. Starting this month, business operators can issue cash receipts and send them to consumers via Son-Tax. Researcher Kim stated, "Fundamental improvements to the current inheritance and gift tax system are necessary to promote SME succession, but it's not easy to implement these changes in the short term due to societal resistance to wealth inheritance." From the 20th, a year-end tax settlement chatbot counseling service will also be provided on mobile. He explained, "For business succession, utilizing a public interest corporation is an option, where the founder sets up a separate corporation pursuing public interest, and this corporation indirectly controls the successor company. This method is used in advanced countries such as the USA and Germany." He continued, "However, in Korea, there are legal restrictions that make it difficult to use due to concerns of inheritance and gift tax evasion and private benefit appropriation by large corporations. The holding limit of donated stocks to a public interest corporation is excessively strict compared to major countries overseas." In the USA, Canada, and Japan, the ownership of voting stocks by a public interest corporation is recognized up to 20\%, 20\%, and 50\% respectively, while in Korea, inheritance and gift taxes are imposed if it exceeds 5-10\%, according to Kim. He emphasized, "The purpose of this regulation is to control irregular inheritance by conglomerates and prevent economic concentration. Applying the same to SMEs is unrealistic. For SMEs eligible for family business inheritance deduction, it's necessary to consider exempting inheritance and gift taxes on the contribution of up to about 20\% of issued voting stocks to a public interest corporation." \\ \hline
\texttt{Image\_caption} & Family Business Inheritanc                                                            \\ \hline
\end{tabular}
\caption{Examples of documents related to specialized domains.}
\label{appendix2}
\end{table*}

\begin{table*}[b]
\begin{tabular}{c|p{12cm}}
\hline
\texttt{Title} &
  Third Nationwide Rally Against Pension Reform in France...Continued Strikes and Traffic Disruptions \\ \hline
\texttt{Subheading} &
  200,000 Participants in Nationwide Protests…Rail and Public Transport Strikes Cause Ongoing Traffic and Logistics Issues, Public Opinion Rebounds with 62\% Supporting the Strike, Prime Minister Philippe Asserts Government and Ruling Party's Firm Commitment to Pension Reform, Labor Unions Remain on Diverging Paths \\ \hline
\texttt{Body\_text} &
  On the 17th, the third nationwide strike against the government's pension reform took place across France. The New York Times reported on the 9th that the prolonged strikes in France are rooted in long-standing class conflicts within French society, stating, "At the heart of France's lengthy strike is a struggle between the haves and the have-nots." In the capital, Paris, protesters gathered at Place de la République and marched to Place de la Nation, and in the afternoon, clashes between protesters and police led to the use of tear gas for dispersion. The newspaper explained that the current conflict between the unions and the government is part of a broader class struggle between the rich and the poor, the privileged and the underprivileged, a conflict that has been present in French society for the past 200 years. On this day, 75.8\% of SNCF engineers and 34\% of railway controllers joined the strike. This strike has historical roots in events such as the French Revolution, which in the late 18th century drove out the privileged nobility and clergy, and the 19th-century conflicts between capitalists and workers. Teachers from various schools also joined the strike, leading to numerous school closures. Philippe Dichban, a sociology professor at the French National Center for Scientific Research, explained that French society is stratified, and the equality pursued by citizens ultimately aims for everyone to be part of the upper class, leading to unending conflicts. The striking parties themselves view this issue not merely as a protest against pension reform but as a struggle between the privileged class and ordinary workers. CFDT, which does not oppose the government's pension reform plan to unify 42 retirement pension systems into one, joined the third rally in opposition to the government's decision to delay retirement age from 62 to 64 years. Philippe Martinez, the leader of the General Confederation of Labor, recently stated in a local broadcast, "There are two visions of social security in opposition," emphasizing, "The essence of pension reform is choosing what kind of society we want." According to AFP, about 200,000 people gathered nationwide for the third rally. On the first rally on the 5th, 800,000 took to the streets nationwide, and on the second rally on the 10th, 339,000 gathered. This perspective is shared by the general union members. Sebastien Proda, a CGT station worker, criticized, "President Macron is only interested in profit-making," adding, "But we are striking not for money, but to provide better services to the public." Although the goal is to redesign the pension system in line with demographic changes due to aging and to reduce the state's financial burden by introducing a single pension system and increasing labor flexibility, the labor sector strongly opposes, arguing, "They want us to work longer and give less pension." They continue, "People in the government come from the world of finance, but we are just fighting for our right to rest after a lifetime of work." The French government is open to negotiating the issue of delayed retirement age, but both sides remain at an impasse. Ahno Buchju, a striking railway engineer, said, "Macron sees everything from the perspective of competition as a financier, while we on strike see it from the perspective of the collective," explaining, "These are completely opposing views."\\ \hline
\texttt{Image\_caption} &
  French Union Members Protesting Against Pension Reform - CGT members holding a strike demonstration in Marseille on the 10th of last month. Scenes of the French Pension Reform Opposition Rally - The fourth general strike in Paris on the 9th. "The French Pension Strike is an Extension of the 18th Century Revolution and the 19th Century Workers vs. Capitalists Struggle" \\ \hline
\end{tabular}
\caption{Examples of documents involving the same Issue.}
\label{appendix3}
\end{table*}

\end{document}